# The Evolving Nature of Latent Spaces: From GANs to Diffusion

Ludovica Schaerf
University of Zurich & Max Planck Society
Culmannstrasse 1, ZH, CH-8006
ludovica.schaerf@uzh.ch

**Abstract**
This paper examines the evolving nature of internal representations in generative visual models, focusing on the conceptual and technical shift from GANs and VAEs to diffusion-based architectures. Drawing on Beatrice Fazi's account of synthesis as the amalgamation of distributed representations, we propose a distinction between "synthesis in a strict sense", where a compact latent space wholly determines the generative process, and "synthesis in a broad sense," which characterizes models whose representational labor is distributed across layers. Through close readings of model architectures and a targeted experimental setup that intervenes in layerwise representations, we show how diffusion models fragment the burden of representation, thereby challenging assumptions about a unified internal space. By situating these findings within media-theoretical frameworks and critically engaging with metaphors such as the latent space and the Platonic Representation Hypothesis, we argue for a reorientation in how generative AI is understood: not as a direct synthesis of content, but as an emergent configuration of specialized processes.

## 1. Introduction

As generative archives (Meyer, 2023), predictive media (Manovich, 2024), stochastic parrots (Bender et al., 2021), or blurry JPEGs of the Web (Chiang, 2023), generative visual models are multifaceted and profoundly influential entities in the contemporary visual sphere. These metaphors, driven by the necessity to grasp the algorithms and their products, connect technical aspects to societal and cultural frameworks. They extricate the current logic of generativism (Martino, 2023) that stems from so-called generative AI, separating the inner predictive logic of these algorithms and their foundation in minimization and statistical methods.

Each of these metaphors carries implicit assumptions about the functioning of the models and their representations. Conversely, these metaphors are at risk of becoming 'folk theories' (Kempton, 1986 as cited in Offert, Dhaliwal, 2024), rather than scientific explanations. In contemporary models, some of these assumptions are only partially fulfilled and become metaphors of metaphors, detached from their technical groundings. An exemplar case of such 'folk theory' is linked to the latent space. Its popularity in the media field arose from experimentation with generative adversarial networks (GANs; Goodfellow et al., 2014) and variational autoencoders (VAEs; Kingma, Welling, 2014) over the last decade. These models, in turn, anchored the definitions of this space to their functionalities:

> "[The latent space] is a multi-dimensional, unperceivable, unimaginable space in which vast quantities of connected images and texts have been encoded (i.e., turned into numerical *vectors*), *compressed* (their number of dimensions has been reduced in order to preserve only some key features), embedded, and *positioned* in relation to one another, according to the statistical frequency of their occurrence together in the various sources from which they are taken" (Somaini, 2024, p.114; own italics)

As we can see from this definition, the latent space is described as an object that summarizes an (input or output) image into a single vector. This vector exhibits distances to other vectors that mirror our perceived distances between the corresponding images, offering a spatial and indexable representation (Schaerf, 2024). Accounts that assign a "quasi-religious role to this space of statistical banality" (Crawford, 2024) are motivated by its vast conceptual affordances, which often feed into metaphors, such as the archive, that were central to media studies discussions in the past half-century (Derrida, 1996; Foster, 2004).

This definition, which reappears with slight variations across the field, implicitly assumes a number of properties within this space. It assumes *density,* as every point within the distribution of this space should be decoded into an image that makes sense (in the context of the training data). The positioning assumes *organization*: two similar images will be projected closer together than dissimilar ones, and images that exhibit a mixture of the two will be placed between them. *Smoothness* should be visible, as the image's surroundings exhibit slight variations in the image itself. Lastly, and most importantly, it assumes that this representation is *relevant* to the output image: that it contributes to a large part (if not all) of its final appearance.

We claim that, even within the short and recent history of generative AI, these models have witnessed a paradigm shift from VAE and GAN models to the current hype surrounding diffusion models (Goodfellow et al., 2014; Kingma, Welling, 2014; Ho et al., 2020), and that this shift problematizes all our implicit assumptions around latent spaces.

We draw from a recent article by Fazi, in which the author discussed the central role of *synthesis* in the deep learning revolution. In the article, synthesis is not intended in the meaning of artifice (i.e., synthetic), but purely "as amalgamation, as a composing and putting together" (Fazi, 2024, p. 3) of representations. The author includes in the definition of synthesis a unity that is a "togetherness of distributed representations" (Fazi, 2024, p. 1), thus accepting cases in which the representations do not come together into a synthetic entity within the model but achieve unity only in relation to others. In this work, we build on her argument but introduce a further distinction between 'synthetic in a strict sense' and 'synthetic in a broad sense'. The former refers to the representations within the model that, taken in isolation, provide the (near) complete representation of the generated output. The latter, instead, refers to representations that retain the full information only if taken holistically and as distributed elements. We claim that this difference, which we will later illustrate with examples, is at the heart of the paradigm shift in the internal representations of vision-generative models.

## 2. Deep Representations

To better understand this distinction, we briefly introduce the concept of representation in deep learning models. In this section, we refer to *deep representations* (or simply representations) as the internal states a model takes on while generating, in our example, an image. This should not be confused with the *weights*, which remain the same across all generations for a given model. The internal states, technically referred to as activations, instead, are the product of the interaction between the model's learned weights and the specific input (in the case of diffusion models, the noise and prompt).

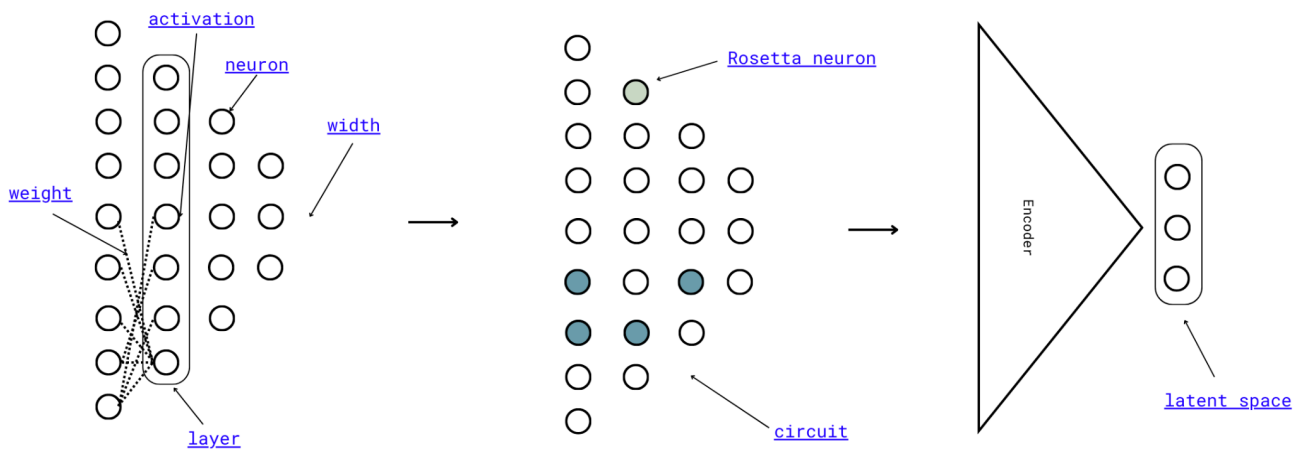

Figure 1: Schematization of a neural network architecture. On the left: introduction basic terminology, some links are omitted for readability. In the center: the neural view, interpretability at the level of the neuron or circuit of neurons. On the right: the layerwise view, the latent space.

In the context of deep learning, advocates for creating and understanding representations have been around since, and possibly even more so, the early days of AI (Rumelhart et al., 1986). Bengio, in a 2016 talk, claims that creating representations within AI models is key to higher-level cognition and abstraction (Bengio, 2016). Broadly speaking, deep representations are created during the model's training, often via backpropagation of errors[1] from the prediction through the weights connecting each layer from the last to the first. The learning proceeds in phases, from utility to compression. An early finding on stochastic gradient descent (SGD), a training optimizer, shows that the learning can be interpreted from an information perspective: neural network models learn to optimize first the representations to perform best in the task and later compress[2] them to the most generalizable configuration (Shwartz-Ziv and Tishby, 2017). Deep representations are distributed (Offert and Impett, 2025). In fact, we do not train individual neurons to represent a specific feature of the output; rather, we let the neurons collectively shape the best possible representation for the task.

Deep representations can be taken at different cardinalities:
- the complete set of activations can be considered one representation,
- the activations at each layer of the model constitute different representations,
- or, the activations of a specific neuron/set of neurons create a representation.

Studies on the latent space belong to the second category; however, the different representational layers interact continuously, and it is important to acknowledge what this choice of focus implies. Deep representation, intended at the neuron's cardinality, is central to interpretability research, which seeks to identify representations of concepts within models to achieve more trustworthy AI (both at the level of single neurons and of neural circuits). This field of study is often closely linked to neuroscience and the representations in the brain. The cardinality of the layer is most commonly studied in the field of representation learning and is frequently used in humanities and social sciences applications and theories. Layerwise[3] representations shift the focus from seeking a specific concept, as in interpretability, to identifying what is represented and how.

The latter's affinity with humanistic approaches becomes apparent: layerwise studies are not interested in specific, delineated concepts; instead, they wish to construct (or apply) theories around incommensurable representations to situate them within a commensurable framework.

---

[1] The errors being backpropagated are called gradients.
[2] For more details on compression in Offert and Impett.
[3] In this case, by layer we mean the output of a collection of neurons operating together at a specific depth within a neural network.

## 2.1. Latent Space

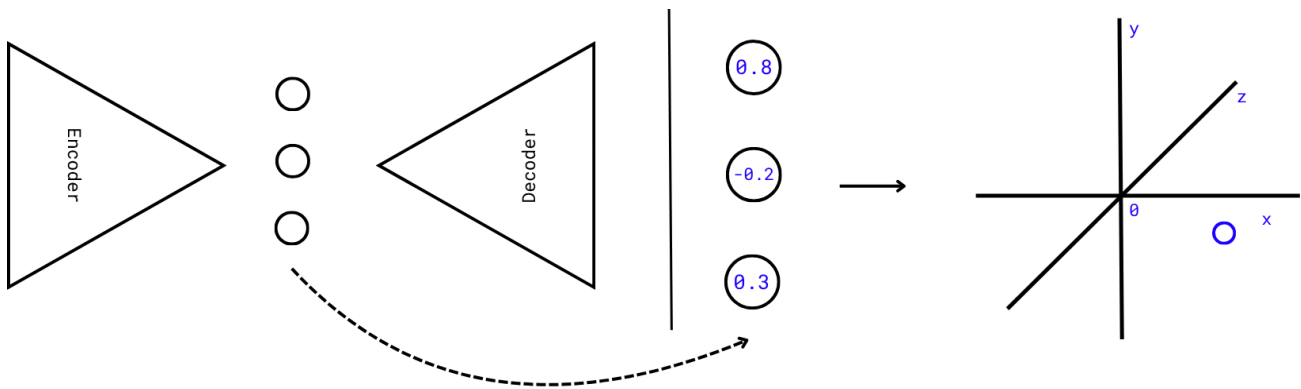

Figure 2: Schematization of an autoencoder with a latent space of three neurons. On the right: example activations of the three neurons and their position in the three dimensional vector space.

The most well-known example of a deep, layerwise representation is the *latent space*. Strictly speaking, the latent space refers to the internal representation formed at the bottleneck layer of a model, where data is compressed into its most compact representation. More broadly, however, the term is often used to describe any layerwise feature representation learned by a neural network.

Consider the figure above. An autoencoder model that compresses the representation of an input image through a series of layers (encoder) into just three nodes — the smallest layer in the model, and therefore called the bottleneck. From there, the representation is expanded to reproduce the original image (decoder). The autoencoder learns by approximating the output image to match the input image. The values at this bottleneck form the latent space vector. Since the bottleneck contains three neurons in one row, these values define a point in a three-dimensional vector space.

The representation in this space is dense, as the bottleneck is highly compressed, forming a continuous space (up to the encoding in a discrete computing machine). A series of techniques and by-products[4] ensures that points in space that are close together yield images we are inclined to consider similar[5]. Often, the representations in the latent space are also independent: not only are similar inputs yielding similar outputs, but moving the input in a specific direction in the space yields a specific variation in the output. In the example above, imagine a direction in the latent space that separates all colored images on one side and all black and white ones on the other. Moving the original point of a black-and-white image in that direction would yield an almost identical color image. These independent directions are common in GANs (most notably StyleGAN) and VAEs (bVAE). Therefore, these representations are dense, smooth, organized, and relevant.

Fazi recalls Chiang's imperfect analogy of ChatGPT to a blurry JPEG of the Web. While there are clear differences underscoring the compression of representations carried out by JPEG and by transformer models (Chiang, 2023; Offert, Impett, 2025), "indirectly, the compression analogy highlights that 1) these technologies are indeed representational; 2) all representing is compressing; and 3) such compressing involves a degree of unification of information." (Fazi, 2024, p. 5)

The third point is particularly significant for our argument: while all representing is compressing, the compression can involve different degrees of unification. In the latent space, the bottleneck

---

[4] The chapter by Schaerf discusses the techniques that determine the continuity and semantic organization of latent spaces.
[5] Similarity is an essential concept in latent space studies, refer to Offert and Impett; Offert for more detailed discussion

layer of the model uniquely and entirely determines the generated output, up to some noise. The unit of re-elaboration and unification of information is contained within this single space and depends solely on the relationships among its elements. Latent spaces of these models create a 'synthesis in a strict sense'.

In fact, the latent space as a place of synthesis has been instrumental to its widespread adoption and theoretical popularity. Nonetheless, as previously mentioned, the recent AI revolution in the visual domain was powered by a shift, around 2020, towards diffusion and autoregressive models. This shift begs the question: do these new models possess latent spaces that are synthetic in the strict sense? What does this imply for the density, smoothness, organization, and relevance of this space?

## 3. Representations Beyond Synthesis

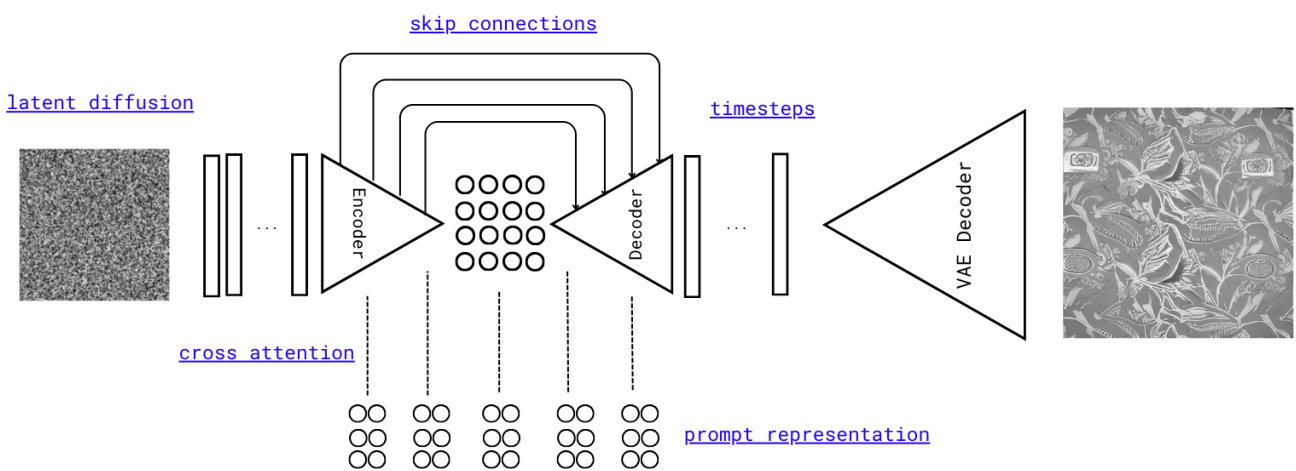

Figure 3: Schematization of a diffusion model with a U-Net backbone. From left to right, it shows: the noise space, the text representations, the different layers and skip connections of the U-Net, the iterative nature, and the latent space before the VAE.

Diffusion models, reintroduced by Ho et al. and later popularized by Stable Diffusion and DALL·E (Rombach et al., 2022; Ramesh et al., 2022), have been metaphorically described as "negentropic" processes (Zylinska, 2024). Rather than reversing entropy in a thermodynamic sense, the metaphor highlights their iterative refinement: they sculpt visual coherence from random noise over time (Salvaggio, 2024). The goal of the model is not to predict the image, but to predict how much noise to subtract at each timestep to carve out the content specified by the prompt. These models exhibit incredible expressivity, but their internal representations are considerably more complex than previous models.

Let us consider now the four assumptions on latent spaces:
1. Density: any point belonging to the latent space distribution outputs in-distribution[6] images
2. Smoothness: moving around a point that generates an in-distribution image yields similar, in-distribution images
3. Organization: the similarity between two points mirrors humanly perceived similarity, and the points in between are mixtures of the former images.
4. Relevance: a change in this space results in notable changes to the output image.

---
[6] Images that are similar to the training images.

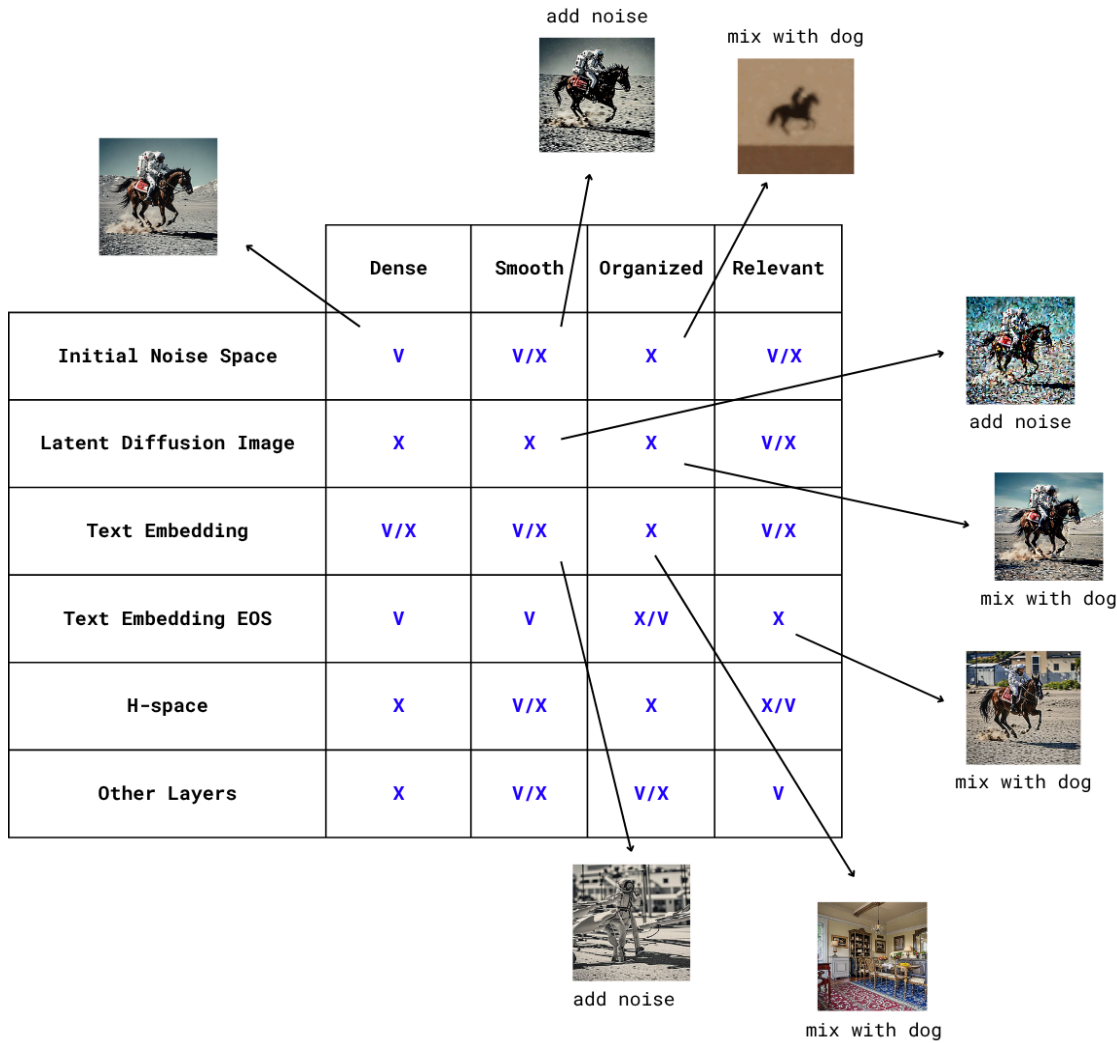

Figure 4: Table summarizing the properties of each candidate layer. We show some examples of adding either noise (in the same distributions as the latent space in question) or the value of the embedding corresponding to 'A dog' in the relevant latent space to substantiate the scoring. V indicates full compliance, while X indicates complete absence.

In diffusion models, the latent space can refer to various components within the architecture. Naturally, it can point to the original noise used in the generation process—the model's input. Latency in this space is intended mainly as a temporal concept. We can test empirically the properties mentioned above. This space is certainly dense, as proven by the fact that you can sample any normally distributed noise and get relevant images. Still, it is only partially smooth and relevant: it has some neighbourhood, but it quickly jumps to a completely different representation. It only affects the image's composition, and it is not organized in any way.

Another natural latent space is the text representation, the other input of the model. In diffusion models, the prompt is represented through a frozen (vision-)language model (such as CLIP or T5; Radford et al., 2021) and incorporated into the model through cross-attention. In this process, each token (roughly a word/subword) is encoded separately, yielding one representation per word. While this is not the diffusion model's latent space, but rather that of the text encoder, it still has a remarkable influence on image generation (relevance). This space is rather dense and smooth (especially the embedding of the End of Sentence, EOS), but it is not visually organized: interpolating between an image of a dog and an astronomer riding a horse yields an interior of a room. Furthermore, the EOS, which is the latent space of the CLIP model, is not relevant to the final image in the diffusion model.

In the branch of diffusion models called latent diffusion, the final denoised image lies in a lower-dimensional space—for instance, 64x64x4 pixels—and is decoded by a VAE into the image space of 256x256 RGB. The lower-dimensional image space is often called the latent space of these models. In practice, this resembles the decompression of a JPEG more closely than our notion of latent spaces: it is neither dense, smooth, nor organized. Its relevance to the final image is purely arbitrary.

This leaves us with only the layerwise representations of the backbone model, predicting the noise at each timestep. For these, we have a separate latent space per timestep because the internal states change at each iteration. Furthermore, diffusion models tend to use internal spaces that are tensor-valued rather than vector-valued: they are groups of spatially organized vectors. Roughly, the bottleneck could consist of 64 vectors distributed in an 8x8 grid, where the top-left vector primarily contains information about the top-left of the image, and so on. Moreover, the information in these models flows through skip connections, i.e., long-range connections between encoder and decoder. This creates a scenario, unlike in previous models, in which not all information passes through the bottleneck (the h-space).

This lack of clarity about which space could serve as the latent space underscores the complexity of these models and prompts us to better understand these representations.

### 3.1: Specialization of Representation

To illustrate the layerwise representations of diffusion models, we conduct the following experiment (Schaerf et al., 2025). We generate two images in parallel; we can call them image O(riginal) and image E(dit). We test what happens when we insert the representation of image E at a specific layer into the generation of image O. We do this for every layer in the model, switching every timestep.

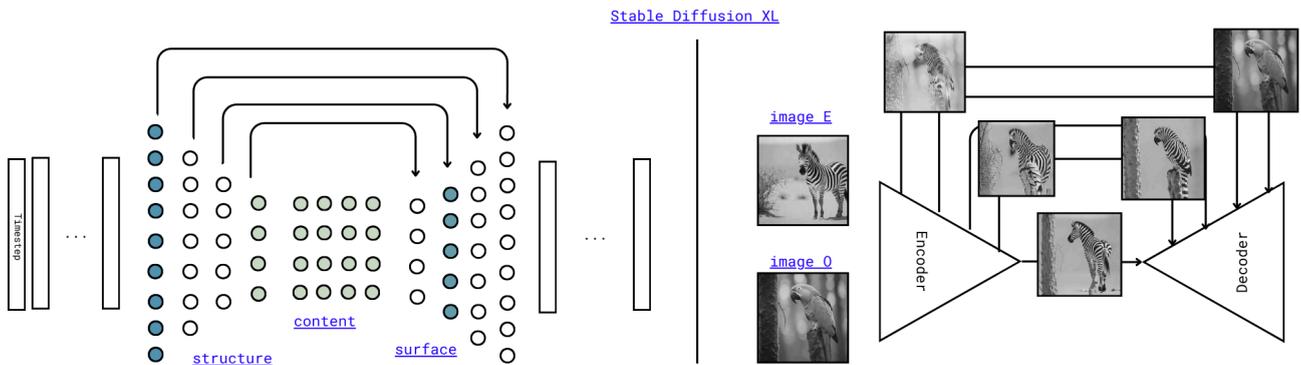

Figure 5: Schematization of the layerwise representations of U-Net diffusion models. On the left: we show which layers are mostly responsible for which aspect of the image. On the right: the results of the injection. We show over each layer the result of injecting that layer of image E into the original image O.

Through this experiment, we identify two main findings[7]. While GANs' and VAEs' latent spaces determine the entire image, diffusion models split the representational burden across layers (split relevance). In every model, we find one or more layers that compose the image *structure*, one or more layers that are responsible for the *content*, and one or more layers that are responsible for

---

[7] The complete discussion of the setup and results of these experiments can be found in Schaerf et al. (2025)

*surface details and colors*. This finding is particularly interesting as it shows a tendency of more complex models with multiple information streams to create a specialization of labour.

We find these three aspects across almost all diffusion models we consider. However, the layers responsible for these aspects are in entirely different parts of the model. Some feature the content layers after the surface layers (i.e., Stable Diffusion 1.4; Rombach et al., 2022), while most have content layers first, and others hold the most critical layers in the bottleneck and early decoder layers (i.e., Stable Diffusion XL; Podell et al., 2023), in contrast to Stable Diffusion 1.4 and 2 that has them in the late encoder. Kandinsky even has all the important layers in the decoder (Razzhigaev et al., 2023). This division of labour can be found even in models that do not rely on the encoder-decoder architecture, but rather on vision transformers (such as Stable Diffusion 3.5 and Flux).

The difference between the internal representations of these models, we believe, problematizes a popular deep learning hypothesis: the Platonic Representation Hypothesis (Huh et al., 2024). This hypothesis argues that many large models converge to similar internal representations due to learning the structure of reality. We contest this by showing that, while representational components may appear similar across models, their internal arrangements remain fundamentally different.

## 4: Combination over Unity

As we have seen, developments in vision models, most notably the iterative process and the introduction of attention modules, remove the need for a unique representation, and the implications of this lead to a loss of density, smoothness, relevance, and organization. They have, however, created a specialization of representations. The information bottleneck of previous models proved too restrictive to achieve the desired expressiveness, leaving room for even more distributed, complex, and iterative representations. We could not find any layer representation in these models that would entirely be responsible for the output, leaving behind a synthesis in a strict sense, but creating a 'synthesis in a broad sense', where the unity of representation lies in the distributed contribution of the several specialized representations in an emergent synthesis.

The Kantian definition of synthetic levels, as mentioned in Fazi, resonates with the specialization of representations found within these models. As Fazi says:

> "He described 1) the synthesis of apprehension in the *intuition*, concerning elemental, raw perceptual inputs and producing temporal and *spatial structures*; 2) the synthesis of reproduction in the *imagination*, a transcendental act of the mind that unifies apprehended representations with other apprehended representations and corresponds to imagination itself; and 3) the synthesis of *recognition in the concept*, which concerns *understanding* and the application of the categories via acts of apperception. What is apprehended in intuition, reproduced in imagination, and recognized in concepts are representations (i.e., intuitions, images, and concepts are representational)." (Fazi, 2024, pp. 6-7)

Kant identifies three faculties of the human mind: intuition (or perception), intellect (or episteme), and reason (or imagination) (Kant, 1998, as cited in Fazi, 2024). We find that, in diffusion models, the first two faculties are reflected in the layerwise representations: some representations are mostly perceptual in nature, with the composition highlighting the spatial structuring, while others are more conceptual, referring to the contents of the images. It is, finally, imagination that aggregates these representations into a coherent and emergent image. Here, imagination is responsible for compositionality, "a concept denoting a whole whose meaning is expressed as a function of the meanings of its combined parts" (Fazi, 2024, p.8).

The connectionist nature of neural networks, whereby meaning is created by the connections between neurons, becomes now even more structural: even the layerwise representations contribute to creating a connectionist representation.